\documentclass[10pt,twocolumn,letterpaper]{article}
\usepackage{arxiv}
\usepackage{times,graphicx,amsmath,amssymb,booktabs,placeins}
\definecolor{eagleblue}{rgb}{0.21,0.49,0.74}
\usepackage[pagebackref,breaklinks,colorlinks,allcolors=eagleblue]{hyperref}
\title{EAGLE: Entity-Aware Grounding with Layered Evidence for High-Resolution Image Understanding}
\author{Zhongjian Zou \quad Quan Qian\\Shanghai University}
\hypersetup{pdfauthor={Zhongjian Zou, Quan Qian},pdftitle={EAGLE: Entity-Aware Grounding with Layered Evidence for High-Resolution Image Understanding}}
\begin{document}
\maketitle
\begin{abstract}
Multimodal large language models (MLLMs) still struggle with fine-grained understanding of high-resolution images: attention heads vary in their reliability for grounding target entities and their attributes, while localized details are often disconnected from the broader scene context. We propose EAGLE (Entity-Aware Grounding with Layered Evidence for High-Resolution Image Understanding), a plug-in framework that jointly selects grounding signals and organizes hierarchical visual evidence. EAGLE employs a fine-grained dataset to screen visually-grounded attention heads, followed by entity-aware grounding to extract target-specific regions. It then integrates hierarchical evidence construction and question-guided routing to select optimal view combinations, feeding them into the frozen MLLM to enhance fine-grained visual understanding. Experiments with Qwen2.5-VL-7B and InternVL3-8B demonstrate that EAGLE consistently improves performance across both backbones and achieves SOTA results on V*, HRBench-4K, and HRBench-8K compared with existing methods using the same backbone.
\end{abstract}

\begin{figure*}[!t]
\centering
\includegraphics[width=0.84\linewidth]{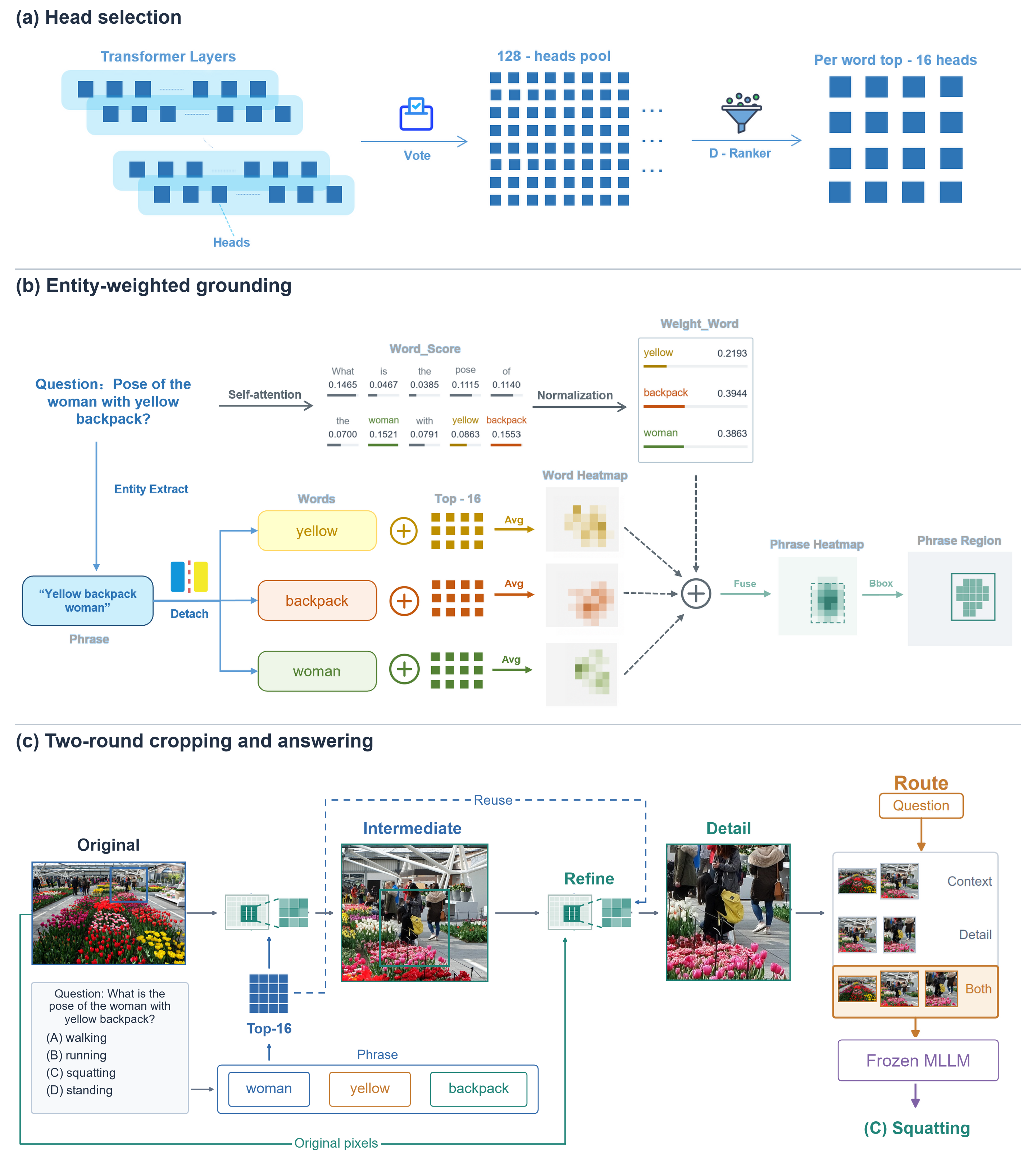}
\caption{EAGLE overview. (a) Offline voting builds a 128-head pool; the dynamic ranker selects Top-K heads per entity word, with K = 16 chosen on the validation set. (b) Question-guided word weights combine their heatmaps into an entity-phrase grounding map. (c) Two localization rounds construct intermediate and detail images from original pixels. Question-guided routing selects the final views for the frozen MLLM.}
\label{fig:1}
\end{figure*}

\section{Introduction}

Multimodal fine-grained recognition is capable of aligning textual entities with visual regions to uncover subtle attributes overlooked by global models, thereby empowering detail-aware decision-making in autonomous driving and smart healthcare. Vision--language pre-training and visual instruction tuning have enabled MLLMs to answer diverse questions about images \cite{ref19,ref20,ref21,ref22,ref23,ref24}. High-resolution images, however, still expose a gap between seeing an image and resolving its fine details \cite{ref3,ref15,ref25}. A small target may occupy only a few visual tokens, while similar objects compete for attention. Even when a crop contains the correct target, the model must connect its appearance to the intended entity and interpret it within the scene. Reliable fine-grained understanding therefore requires both accurate grounding and suitable visual context.

Prior crop-based methods recover local evidence through visual search or attention-guided region extraction \cite{ref3,ref4,ref5,ref16,ref17}. However, previous token-wise localization methods can retain responses to non-localization words and combine regions that refer to different objects. For ``woman in black jacket'', independently localizing ``black'', ``jacket'', and ``woman'' may preserve unrelated objects rather than their joint referent. EAGLE removes non-localization words such as ``in'' and fuses the remaining word heatmaps within the entity phrase before region extraction. Figure~\ref{fig:2} contrasts this phrase-level grounding with token-wise localization.

Selecting visual grounding heads is another challenge. Prior studies show that only a subset of attention heads consistently supports visual localization \cite{ref6,ref7}. Heads without this capability often produce diffuse or unrelated image responses, which act as noise when used for grounding. The key is therefore to identify suitable visual heads before combining their heatmaps. Available data also vary in fine-grained difficulty: in the source examples we inspected, some target objects occupy more than 30\% of the original image, making them less informative for screening heads that must locate small targets. We curate and manually annotate Visual-CoT (fine-grained) to build a specialized visual-head pool, then use a trained ranker to select the heads most suitable for the current entity.

Correct localization does not guarantee a correct answer. Even a crop that clearly exposes the target can fail to provide sufficient evidence for recognition. Prior original-plus-crop pipelines \cite{ref4,ref5} leave two related difficulties. A tight crop removes context needed to identify the target, while the large scale gap between the original image and the magnified detail can hinder cross-view association. Contextual blindness has also motivated multi-scale cropping \cite{ref9}. EAGLE retains an intermediate view to connect target details with their surroundings, then selects the view combination according to the question. The examples in Figure~\ref{fig:4} show that adding detail alone does not consistently improve predictions.

We present EAGLE, a unified framework connecting entity-aware grounding, layered evidence construction, and question-guided routing (Figure~\ref{fig:1}). We curate Visual-CoT (fine-grained) from Visual CoT \cite{ref11} and manually annotate fine-grained entity regions to build a specialized candidate head pool. A trained ranker selects the Top-K heads for the current image and entity word. Question-guided word weights fuse their maps within each entity. The resulting regions form an intermediate image for further localization and contextual reasoning. After removing invalid padding regions, refined targets are sampled from the original pixels to form a detail image. The router then chooses the image combination needed to answer the question.

Our contributions are as follows:

\begin{itemize}

\item To address non-semantic noise interference in entity grounding, we adopt a phrase-level grounding approach. It eliminates non-localization word noise and uses question self-attention to weigh and fuse phrase heatmaps. 

\item To eliminate the influence of non-localizable heads in Transformer architectures, we employ a voting mechanism on our self-annotated fine-grained dataset. This mechanism successfully screens out fine-grained visual attention heads, thereby yielding high-quality grounding heatmaps.

\item To enrich the context of detail crops and bridge the scale gap between global scenes and local crops, we construct layered evidence and utilize question-guided routing to select complementary view combinations, facilitating cross-scale evidence association.

\item Among same-backbone methods utilizing Qwen2.5-VL-7B and InternVL3-8B as backbone MLLMs, our method achieves SOTA performance on all three benchmarks: V*, HRBench-4K and HRBench-8K.

\end{itemize}

\section{Related work}

\paragraph{Previous studies on multimodal visual understanding and alignment.} Vision Transformers represent images as patch sequences \cite{ref27}, while CLIP learns transferable visual representations from image--text pairs \cite{ref19}. Flamingo connects visual encoders with language models for few-shot multimodal learning \cite{ref20}, and BLIP-2 uses a lightweight querying module to bridge frozen visual and language models \cite{ref21}. LLaVA and InstructBLIP establish visual instruction tuning as a route to general-purpose image conversation \cite{ref22,ref23}; LLaVA-1.5 improves this recipe with a simple connector and task-oriented data \cite{ref24}. These advances establish the backbone capabilities on which EAGLE builds. Our focus is how an existing model can access and combine the evidence needed for a fine-grained question.

\paragraph{Fine-grained perception on high-resolution images.} Early MLLMs commonly encoded images at a fixed resolution, limiting access to small objects and text. LLaVA-UHD addresses varying resolutions and aspect ratios through image slicing, token compression, and spatial organization \cite{ref25}. InternVL scales visual representation learning \cite{ref26}, and PS3 introduces selective local pre-training for 4K inputs \cite{ref8}. Qwen2.5-VL and InternVL3 further strengthen high-resolution perception \cite{ref1,ref2}. Complementing these backbone advances, V* uses guided visual search \cite{ref3}; ViCrop, ZoomEye, and retrieval-augmented perception recover useful regions at inference \cite{ref4,ref16,ref17}. EAGLE also operates at inference, but jointly selects entity-specific grounding signals and the views supplied for answering.

\paragraph{Attention-based heatmap grounding.} Attention rollout traces information across Transformer layers \cite{ref28}, while recent work shows that individual MLLM heads differ in visual perception and grounding ability \cite{ref6,ref7}. ViCrop and HiDe both employ attention-based heatmap localization \cite{ref4,ref5}. HAVC combines offline OCR-based head screening with online heatmap selection \cite{ref10}. In contrast to these generic approaches, we construct Visual-CoT, a manually annotated dataset, to identify a specialized head pool and dynamically select heads tailored to the specific target.

\section{Method}

\begin{figure*}[!t]
\centering
\includegraphics[width=1.0\linewidth]{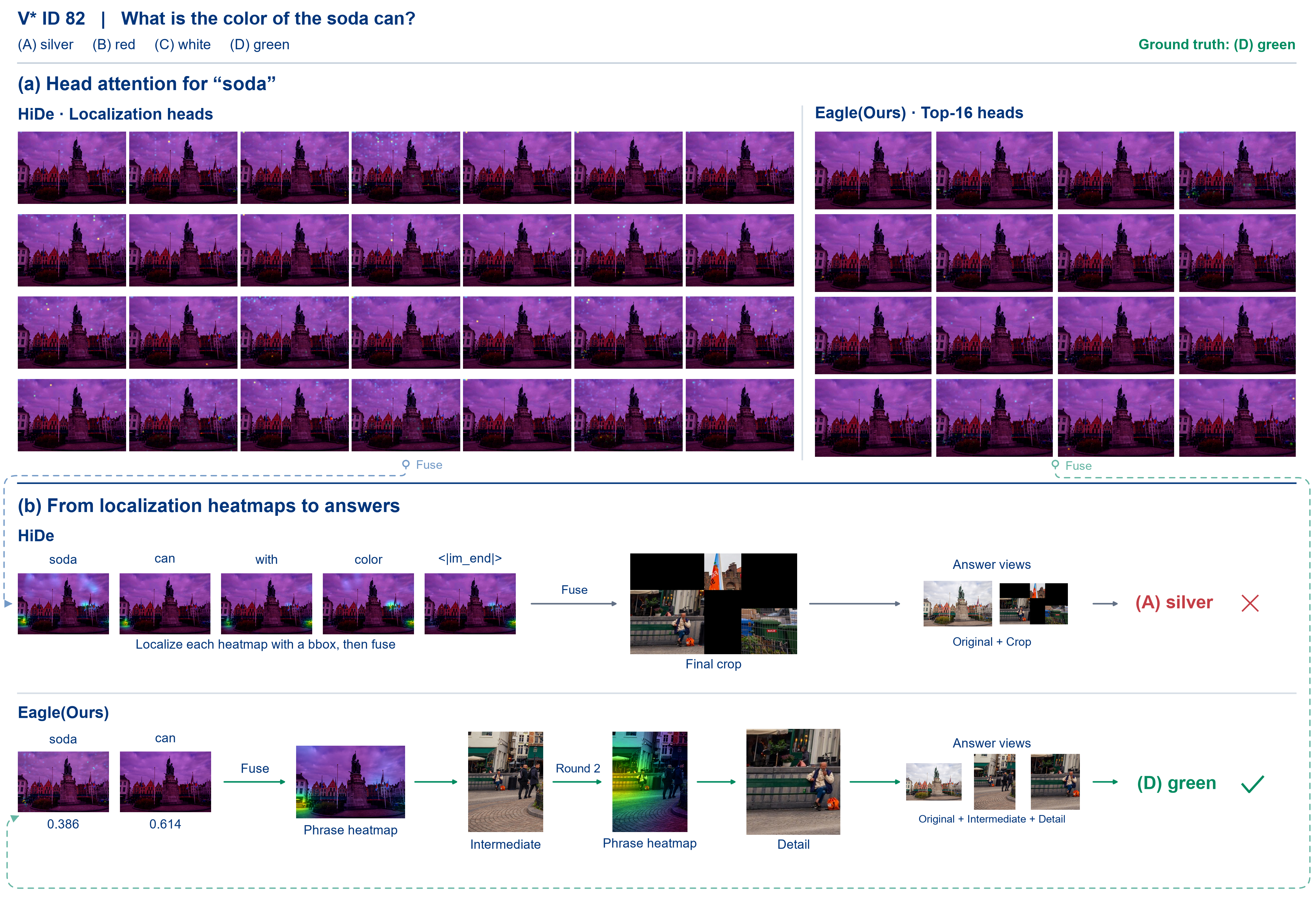}
\caption{Head selection and entity grounding on the soda-can example. (a) Attention maps for ``soda'' from all 28 heads of the displayed HiDe layer and EAGLE's selected Top-16 heads. (b) HiDe's token-wise crops lead to ``silver'', whereas EAGLE combines ``soda'' and ``can'', refines the region, and answers ``green''. }
\label{fig:2}
\end{figure*}

\subsection{Overview and entity extraction}

Given an image I and a question Q, EAGLE identifies the target entities, constructs informative views, and passes the selected views to a frozen MLLM for answering (Figure~\ref{fig:1}). Candidate-head screening and ranker learning are performed offline. Online selection uses the current image and extracted entity words, without access to reference answers or target bounding boxes.

Entity extraction removes words that do not identify visual content and retains the object, part, and modifiers needed to specify the target. For ``woman in black jacket'', it removes ``in'' and forms ``black jacket woman''. For a relation between a woman with a backpack and a person with a beanie, the two entity phrases remain separate. We fuse the word heatmaps within each phrase before extracting regions, rather than locating every word separately and merging all its boxes. This preserves a phrase as one target and limits unrelated regions activated by individual words.

The extracted words are mapped to their positions in the question so that their importance can be estimated consistently. Regions from different entities are retained together when constructing the intermediate image. Thus, a relational question can preserve both the queried object and its reference object rather than enlarging one while losing the other.

\subsection{Fine-grained head pool and dynamic ranking}

\paragraph{Building the head pool.} We construct Visual-CoT (fine-grained) from Visual CoT \cite{ref11} and manually link entity phrases to fine-grained image regions. Candidate heads are evaluated against these annotations. Let G denote the annotated target region, R the retained region, and $\Omega$ the full image. We measure target coverage and background removal (BR) as:

\begin{equation}\begin{aligned}
\operatorname{Coverage}(R,G)&=\frac{|R\cap G|}{|G|},\\
\operatorname{BR}(R,G)&=1-\frac{|R|-|R\cap G|}{|\Omega|-|G|}.
\end{aligned}\label{eq:1}\end{equation}

Only heads with target coverage of at least 0.90 and background removal of at least 0.50 are eligible for voting. Each entity votes for up to ten eligible heads, favoring those that cover the target and concentrate attention inside it.

Votes are accumulated across the annotated entity--image pairs, and the 128 highest-ranked heads form the candidate pool C. Ties are resolved by aggregate localization quality. This step identifies visual heads suited to fine-grained grounding. Online ranking then selects which of these heads best responds to the current image and entity word.

\paragraph{Ranking current heatmaps.} Visual grounding ability is not uniform across heads, prior analyses identify context-dependent visual activation \cite{ref7} and a small subset of heads suited to localization \cite{ref6}. These findings motivate selecting heads for the current entity rather than treating all visual heads as interchangeable. The ranker uses eight heatmap features: total attention, component-area entropy, edge alignment, top-three component mass, fragmentation, largest-component mass, peak component response, and the mass of the component containing the global peak. After Min--Max normalization within the pool, a learned linear score selects the Top-K heads for equal-weight map fusion (Figure~\ref{fig:1}a):

\begin{equation}\begin{aligned}
s_{i,h}&=\boldsymbol\theta^{\top}\widehat{\mathbf f}_{i,h},\qquad
S_i=\operatorname{TopK}_{h\in\mathcal C}s_{i,h},\\
H_i&=\frac{1}{K}\sum_{h\in S_i}A_{i,h}.
\end{aligned}\label{eq:2}\end{equation}

Here $\mathbf{f}_{i,h}$ describes the heatmap of head h for word i, $\boldsymbol{\theta}$ contains the learned ranking weights, and $A_{i,h}$ is the word-to-image attention map. We select K on the validation set and use K = 16 in all experiments. K remains fixed at inference, while the selected head identities change with the input. Ranking scores determine which heads are selected; their maps are averaged equally. Aggregating several ranked heads limits the influence of an occasional unreliable head on the fused map. The second localization pass reuses these heads and the word weights, but recomputes attention on the intermediate image.

\paragraph{Learning the ranker.} Manual target annotations order heatmaps for the same entity word by target containment, then recall-oriented overlap (F2), coverage, and intersection over union (IoU). A preferred and a less preferred candidate form a pair $(h^{+},h^{-})$, with feature difference $\Delta\mathbf{f}=\mathbf{f}^{+}-\mathbf{f}^{-}$. The pairwise loss encourages the preferred candidate to receive a higher score; Figure~\ref{fig:3} evaluates the resulting selection against random sampling from the same pool:

\begin{equation}
\mathcal L_{\mathrm{rank}}=\frac{1}{|\mathcal P|}\sum_{(h^+,h^-)\in\mathcal P}
\log\!\left(1+\exp(-\boldsymbol\theta^\top\Delta\mathbf f)\right).
\label{eq:3}\end{equation}

P is the set of ordered training pairs; tied candidates are excluded. The ranker is trained offline and frozen during inference. Only this small auxiliary model is learned; the MLLM parameters remain unchanged. Thus, EAGLE adapts the choice of grounding signals without fine-tuning the backbone.

\subsection{Question-guided entity heatmaps}

\paragraph{Question-guided word weights.} Words within an entity phrase contribute unequally to a question. Self-attention provides word-importance scores, following attention-based attribution \cite{ref28}. Normalizing the scores within each phrase gives word weights $\alpha_i$; a single-word entity receives weight one. These weights combine the selected-head maps in Eq.~\eqref{eq:4}, emphasizing question-relevant cues. Figure~\ref{fig:1}b illustrates the progression from word scores to a fused phrase heatmap.

Shared-region enhancement applies when an entity phrase contains at least two words. Let m denote the number of words in that phrase ($m\geq 2$), and let $\widetilde H_i$ be the spatially normalized heatmap of its i-th word. Averaging the pixelwise minimum over all word pairs yields the shared response $C_e$. The final entity map combines the weighted word maps with this response:

\begin{equation}\begin{aligned}
C_e(p)&=\frac{2}{m(m-1)}\sum_{i<j}\min\!\left(\widetilde H_i(p),\widetilde H_j(p)\right),\\
P_e(p)&=\sum_{i\in e}\alpha_i\widetilde H_i(p)+\mathrm{PC}\,C_e(p).
\end{aligned}\label{eq:4}\end{equation}

The shared term reinforces agreement without imposing a hard intersection. For ``white cup'', it favors locations supported by both ``white'' and ``cup'', while the weighted base map retains complementary evidence. PC is the positive coefficient controlling this enhancement, and p denotes an image location in Eq.~\eqref{eq:4}. Single-word entities use their own map. Smoothing, normalization, and thresholding yield the retained regions. Figure~\ref{fig:2} shows the complete localization-to-answer pipeline; its improvement reflects the combined grounding and evidence construction, rather than isolating head selection alone.

\begin{table*}[!t]
\caption{High-resolution question answering accuracy (\%). $\Delta$ reports gains over Base in percentage points, computed before rounding. Bold marks within-backbone column maxima, including rounded ties. EviSpec \cite{ref18} is a concurrent preprint.}
\label{tab:1}
\centering\small
\setlength{\tabcolsep}{4pt}
\begin{tabular}{lrrrrrrrrr}
\toprule
& \multicolumn{3}{c}{V*Bench} & \multicolumn{3}{c}{HRBench-4K} & \multicolumn{3}{c}{HRBench-8K} \\
\cmidrule(lr){2-4}\cmidrule(lr){5-7}\cmidrule(lr){8-10}
Method & Attr & Spatial & Avg & FSP & FCP & Avg & FSP & FCP & Avg \\
\midrule
\multicolumn{10}{l}{\textbf{Qwen2.5-VL-7B}} \\
Base & 80.0 & 76.3 & 78.5 & 89.0 & 56.0 & 72.5 & 83.2 & 53.2 & 68.2 \\
ViCrop \cite{ref4} & 89.6 & 71.1 & 82.2 & 90.5 & 57.5 & 74.0 & 85.5 & 53.0 & 69.3 \\
DeepEyes \cite{ref13} & 91.3 & 88.2 & 90.1 & 91.3 & 59.0 & 75.1 & 86.8 & 58.5 & 72.6 \\
HiDe \cite{ref5} & 92.2 & 89.5 & 91.1 & 95.3 & 58.3 & 76.8 & 93.8 & 58.8 & 76.3 \\
DeepScan \cite{ref12} & 93.0 & 86.8 & 90.6 & 90.1 & 59.7 & 75.0 & 87.2 & 57.6 & 72.4 \\
EviSpec \cite{ref18} & \textbf{95.7} & 86.8 & 92.1 & \textbf{97.0} & 64.8 & 80.9 & \textbf{97.5} & 65.8 & 81.6 \\
EAGLE & 94.8 & \textbf{94.7} & \textbf{94.8} & 94.3 & \textbf{67.8} & \textbf{81.0} & 94.8 & \textbf{69.0} & \textbf{81.9} \\
$\Delta$ vs. Base & +14.8 & +18.4 & +16.3 & +5.3 & +11.8 & +8.5 & +11.6 & +15.8 & +13.7 \\
\midrule
\multicolumn{10}{l}{\textbf{InternVL3-8B}} \\
Base & 82.6 & 76.3 & 80.1 & 82.0 & 56.8 & 69.4 & 80.5 & 57.0 & 68.8 \\
ViCrop \cite{ref4} & 88.7 & 75.0 & 83.3 & 88.0 & 57.0 & 72.5 & 82.8 & 54.8 & 68.8 \\
HiDe \cite{ref5} & 92.2 & \textbf{90.8} & 91.6 & 90.0 & 63.5 & 76.8 & 92.2 & \textbf{62.0} & 77.1 \\
EviSpec \cite{ref18} & 90.4 & 88.2 & 89.5 & 93.8 & 61.0 & 77.4 & 90.8 & 59.8 & 75.3 \\
EAGLE & \textbf{95.7} & \textbf{90.8} & \textbf{93.7} & \textbf{94.3} & \textbf{63.8} & \textbf{79.0} & \textbf{92.8} & \textbf{62.0} & \textbf{77.4} \\
$\Delta$ vs. Base & +13.1 & +14.5 & +13.6 & +12.3 & +7.0 & +9.6 & +12.3 & +5.0 & +8.6 \\
\bottomrule\end{tabular}
\end{table*}

\subsection{Layered evidence and question-guided routing}

\paragraph{Constructing intermediate and detail images.} Figure~\ref{fig:1}c shows two localization rounds. Eagle uses HiDe's layout-preserving reorganization \cite{ref5} to remove irrelevant rows and columns while retaining the arrangement of selected regions, producing the intermediate image $I_m$. A recorded coordinate mapping connects $I_m$ to the original image I. The second pass localizes targets within $I_m$, removes geometrically invalid padding, and maps the refined regions back to I. Sampling the original pixels produces the detail image $I_d$.

The two constructed images have complementary objectives. Because the detail image depends on the intermediate image, the first pass favors retaining as many relevant regions as possible, preserving both target evidence and useful context. The intermediate image thus provides the search space for refinement and bridges the original scene and the detail image. The second pass instead favors stronger removal of irrelevant background so that local target features become more prominent. Both views are reconstructed from original pixels; invalid padding is removed geometrically rather than by color. The original image is retained when complete scene coverage or distance information is required.

\paragraph{Question-guided routing.} Different questions require different visual evidence. The router selects the original and intermediate images for broad layout or contextual interpretation, such as a sign's meaning or an object's function; the intermediate and detail images for clearly localized attributes, text, faces, and other fine details; and all three images when both scene-level information and local details are needed, including full-image counting and near--far comparisons. The decision follows the information required by the question, so the same object can use different routes under different questions.

\section{Experiments}

\begin{figure}[!t]
\centering
\includegraphics[width=1.0\linewidth]{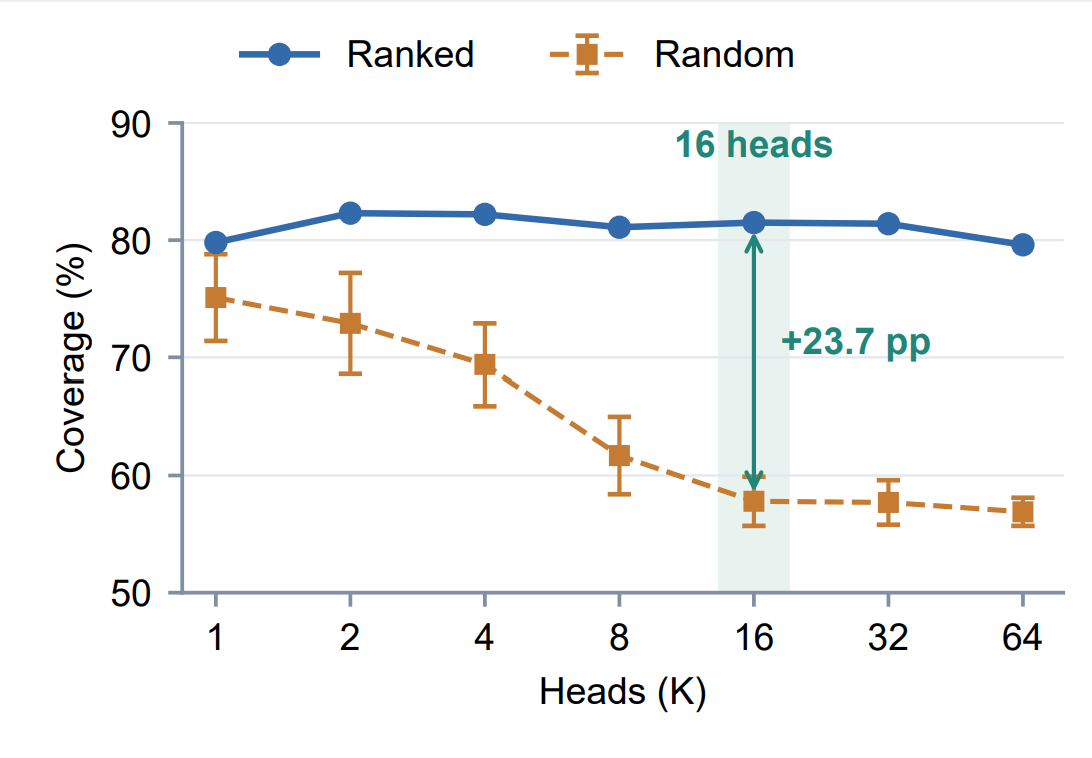}
\caption{Ranked versus random selection within the same 128-head pool. Random selection is repeated ten times; error bars show one standard deviation. At K = 16, the ranker achieves 81.5\% coverage versus 57.8\% for random selection, a gain of 23.7 percentage points.}
\label{fig:3}
\end{figure}

\begin{table}[!t]
\caption{Component ablations with Qwen2.5-VL-7B. Accuracy (\%). Attr, Spatial, and Avg denote attribute, spatial, and overall accuracy. O, I, and D denote the original, intermediate, and detail images. Fixed-view variants disable routing. Bold marks the best result in each column.}
\label{tab:2}
\centering\small
\setlength{\tabcolsep}{3pt}
\begin{tabular}{lrrr}
\toprule
Method & Attr & Spatial & Avg \\
\midrule
w/o Entity phrase extraction & 90.43 & 88.16 & 89.53 \\
w/o Head selection & 91.30 & 85.53 & 89.01 \\
Fixed views: O + I & 90.43 & 92.11 & 91.10 \\
Fixed views: I + D & 93.04 & 92.11 & 92.67 \\
Fixed views: O + I + D & 93.91 & 92.11 & 93.19 \\
\midrule
\textbf{EAGLE (Full)} & \textbf{94.78} & \textbf{94.74} & \textbf{94.76} \\
\bottomrule\end{tabular}
\end{table}

\begin{figure*}[!t]
\centering
\includegraphics[width=1.0\linewidth]{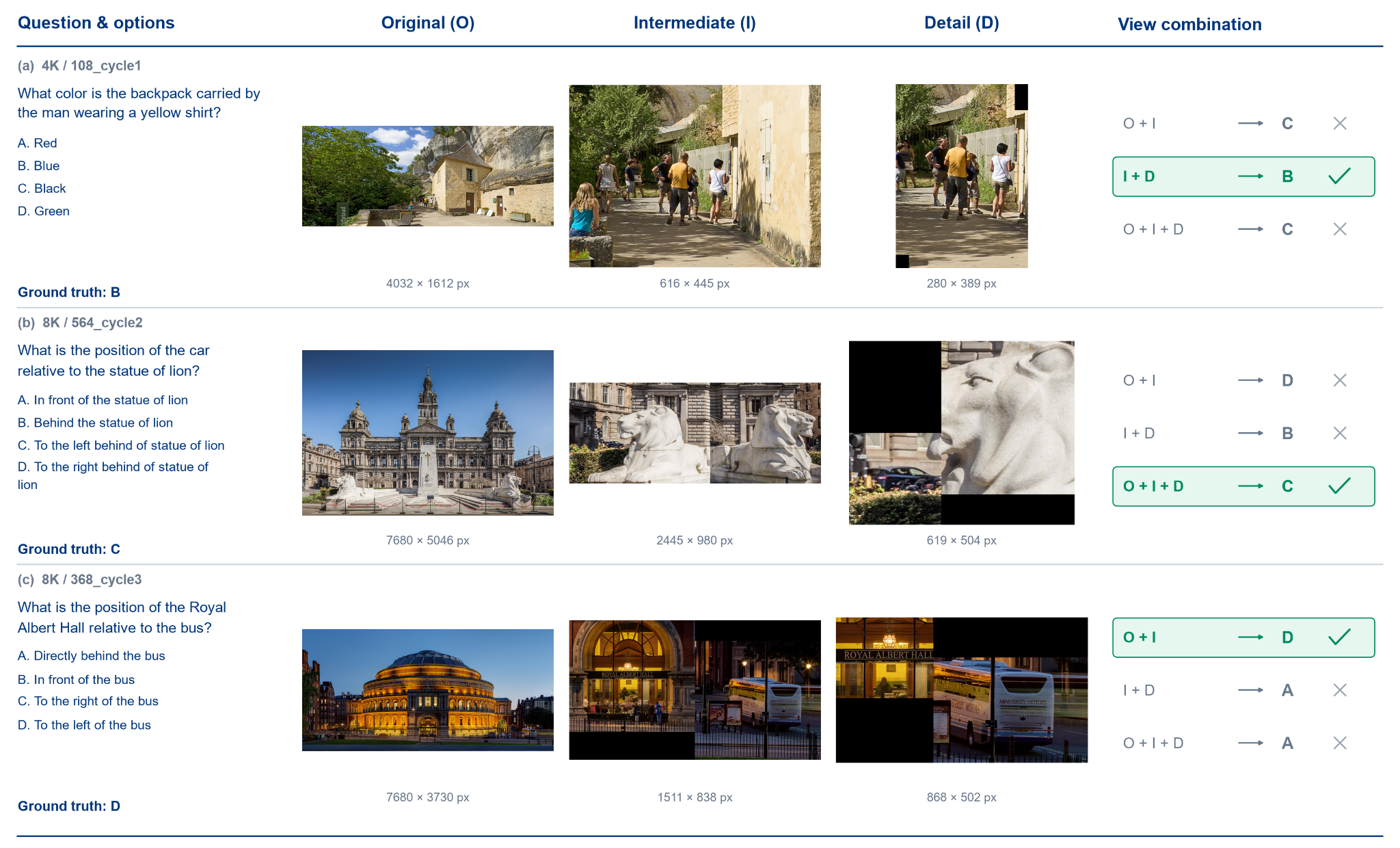}
\caption{View-combination examples. O, I, and D denote the original, intermediate, and detail views. Only I + D correctly identifies the blue backpack in (a); O + I + D recovers the car--lion relation in (b); and O + I preserves the context needed for the hall--bus relation in (c). Green marks the correct prediction for each case.}
\label{fig:4}
\end{figure*}

\subsection{Experimental setup}

\paragraph{Benchmarks and backbones.} We evaluate EAGLE with Qwen2.5-VL-7B and InternVL3-8B on V*Bench, HRBench-4K, and HRBench-8K \cite{ref1,ref2,ref3,ref14}. V*Bench contains 191 questions: 115 attribute questions and 76 relative-position questions. Each HRBench version contains 800 evaluation questions: 400 for fine-grained single-instance perception (FSP) and 400 for fine-grained cross-instance perception (FCP), counting the released option-order variants. We report subgroup and overall accuracy and compare methods within the same backbone family.

\paragraph{Fine-grained supervision.} We curate Visual-CoT (fine-grained) from Visual CoT \cite{ref11}, select examples that require fine-grained entity localization, and manually annotate their target regions. These entity--image annotations support candidate-head voting and ranker learning. A separate localization set, excluded from voting and ranker fitting, evaluates head selection. This procedure identifies useful visual heads offline; online inference does not require target annotations.

\paragraph{Hardware and software environment.} The experimental server is equipped with four NVIDIA H20 GPUs, each reporting 97,871 MiB of memory (approximately 95.6 GiB). The software environment consists of Linux (kernel 5.15.0-139-generic), Python 3.11.4, PyTorch 2.6.0 with CUDA 12.4, Transformers 4.57.1, and NVIDIA driver 550.163.01.

\subsection{Evaluation on high-resolution fine-grained benchmarks}

To evaluate EAGLE on high-resolution fine-grained visual question answering, we compare its performance on V*Bench, HRBench-4K, and HRBench-8K with that of methods using the same backbone. Table~\ref{tab:1} compares EAGLE with existing methods within each backbone family. EAGLE achieves the highest overall accuracy among the listed methods on all three benchmarks: 94.8\%, 81.0\%, and 81.9\% with Qwen2.5-VL-7B, and 93.7\%, 79.0\%, and 77.4\% with InternVL3-8B. Gains over the corresponding backbone baselines range from 8.5 to 16.3 percentage points. Consistent improvements across two model families and three benchmarks provide evidence of cross-model generalizability within the evaluated setting.

The subgroup results in Table~\ref{tab:1} further show strong performance on instance relations. Qwen2.5-VL-7B reaches 94.7\% on spatial questions and 67.8\%/69.0\% on HRBench-4K/8K FCP. Relative to the concurrent EviSpec preprint \cite{ref18}, its overall scores are higher by 2.7, 0.1, and 0.3 points. With InternVL3-8B, EAGLE exceeds HiDe by 2.1, 2.2, and 0.3 points overall.

\subsection{Effect of the ranking function}

To assess whether the learned ranking function selects attention heads that better cover the target region, we compare ranked and random head selection under otherwise identical settings. Figure~\ref{fig:3} compares ranked and random head selection on the held-out localization set. Both use the same 128-head pool, map fusion, and region extraction. Random selection is repeated ten times for each K, with heads sampled without replacement. The standard deviation measures variation caused by random selection.

Figure~\ref{fig:3} shows that ranked selection maintains high coverage across K, whereas random selection deteriorates as more heads are averaged. At K = 16, coverage is 81.5\% versus 57.8\%, a 23.7-point gain. K = 2 and K = 4 achieve slightly higher coverage, so this plot does not establish K = 16 as the coverage optimum. The adopted K = 16 aggregates more candidate signals while limiting each head to 1/16 of the fused map.

\subsection{View combinations and the need for routing}

To assess the contributions of the grounding modules and determine whether question-guided routing improves on fixed view combinations, we conduct module ablations and compare alternative view configurations. Table~\ref{tab:2} separates the grounding components from view selection. Removing the entity-phrase design reduces accuracy from 94.76\% to 89.53\% (5.24 percentage points before rounding); this variant jointly replaces phrase extraction with baseline extraction, uses equal word weights, and disables shared-region enhancement. Removing head selection yields 89.01\% (a 5.76-point drop). These are module-level ablations, not isolated tests of each subcomponent. With grounding and answering settings fixed, O + I, I + D, and O + I + D reach 91.10\%, 92.67\%, and 93.19\%, respectively. The routed method outperforms all three fixed combinations, supporting question-dependent evidence selection.

To examine how the visual evidence needed for answering varies across questions, we qualitatively compare predictions obtained with different view combinations. Figure~\ref{fig:4} illustrates how the useful view combination depends on the question. Intermediate and detail views reveal the backpack color, all three views support the car--lion relation, and original plus intermediate views preserve the context for the hall--bus relation. In these examples, adding or removing a view changes the prediction even though the target scene is unchanged.

\section{Conclusion}

EAGLE combines entity-aware grounding with layered visual evidence for high-resolution understanding. Removing non-localization words and fusing phrase-level responses improves target specification; dynamic Top-16 selection supplies entity-specific grounding signals. The newly introduced intermediate views provide detailed context and bridge the scale gap between global and local images, while question-guided routing is responsible for selecting suitable visual evidence for answering. Results on three benchmarks with two frozen backbones support the framework's cross-model applicability and achieve SOTA performance among the evaluated same-backbone methods.

Future work will evaluate EAGLE on a broader range of multimodal large language models to assess its generalizability beyond the two backbones studied here. To address the limited availability of high-quality, openly accessible fine-grained datasets, we also plan to construct datasets with more detailed category annotations and fine-grained visual questions, supporting more comprehensive evaluation and further research in this area.

\FloatBarrier

{\small\bibliographystyle{ieeenat_fullname}\bibliography{references}}
\end{document}